\documentclass{llncs}
\pdfoutput=1 %arXiv admin added this line
\usepackage{times}
\usepackage{epic}
\usepackage{graphicx}
\usepackage{latexsym}
\usepackage{amsmath}
\usepackage{verbatim}
\usepackage{xspace}

\begin{document}

\input epsf

%\usepackage{amssymb}

%\addtolength{\textwidth}{1.5in}
%\addtolength{\textwidth}{0.5in}
%\addtolength{\oddsidemargin}{-0.75in}
%\addtolength{\topmargin}{-0.5in}
%\addtolength{\textheight}{0.5in}

%\newcommand{\figref}[1]{Figure \ref{#1}}
%\newcommand{\tblref}[1]{Table \ref{#1}}
%\newcommand{\secref}[1]{Section \ref{#1}}
%\newcommand{\eqref}[1]{(\ref{#1})}
%\newcommand{\mygtrsim}{\gtrsim}

%\newtheorem{theorem}{Theorem}
%\newtheorem{definition}{Definition}
%\newtheorem{example}{Example}
%\newtheorem{theorem1}{Theorem}
%\newcommand{\proof}{\noindent {\bf Proof:\ \ }}
%\newcommand{\qed}{\mbox{$\Box$}}
%\newcommand{\qed}{\mbox{QED.}}

\newcommand{\set}{\mathcal}
\newcommand{\myset}[1]{\ensuremath{\mathcal #1}}

\renewcommand{\theenumii}{\alph{enumii}}
\renewcommand{\theenumiii}{\roman{enumiii}}
\newcommand{\figref}[1]{Figure \ref{#1}}
\newcommand{\tref}[1]{Table \ref{#1}}
\renewcommand{\And}{\wedge}
\newcommand{\myldots}{\ldots}

\newtheorem{mydefinition}{Definition}
\newtheorem{mytheorem}{Theorem}
\newtheorem{mylemma}{Lemma}
\spnewtheorem*{myexample}{Running Example}{\bf}{\it}
\spnewtheorem*{myexample2}{Example}{\bf}{\it}
\newtheorem{mytheorem1}{Theorem}
\newcommand{\myproof}{\noindent {\bf Proof:\ \ }}
\newcommand{\myqed}{\mbox{$\Box$}}

\newcommand{\constraint}[1]{\mbox{\sc #1}}
\newcommand{\alldiff}{\constraint{All-Different}\xspace}
\newcommand{\regular}{\constraint{Regular}\xspace}
\newcommand{\grammar}{\constraint{Cfg}\xspace}
\newcommand{\alldiffandsum}{\constraint{All-Different+Sum}\xspace}
\newcommand{\gcc}{\constraint{GCC}\xspace}
\newcommand{\egcc}{\constraint{eGCC}\xspace}
\newcommand{\costgcc}{\constraint{Cost-GCC}}
\newcommand{\softgcc}{\constraint{SoftGCC}\xspace}
\newcommand{\minmax}{\constraint{MinMax}}
\newcommand{\dom}{\ensuremath{\mbox{dom}}}
\newcommand{\SwitchC}{\constraint{Switch}}
\newcommand{\MaxSwitch}{\constraint{MaxSwitch}}
\newcommand{\MinSwitch}{\constraint{MinSwitch}}
\newcommand{\sumc}{\constraint{Sum}\xspace}
\newcommand{\interdistance}{\constraint{Inter-Distance}}
\newcommand{\nvalue}{\constraint{NValue}\xspace}
\newcommand{\AtMostNValue}{\constraint{AtMostNValue}\xspace}
\newcommand{\AtLeastNValue}{\constraint{AtLeastNValue}\xspace}
\newcommand{\permutation}{\constraint{Permutation}\xspace}
\newcommand{\mypermutation}{\constraint{Same}\xspace}
\newcommand{\sameset}{\constraint{SameSet}\xspace}
\newcommand{\same}{\constraint{Same}\xspace}
\newcommand{\usedby}{\constraint{Used-By}\xspace}
\newcommand{\uses}{\constraint{Uses}\xspace}
\newcommand{\disjoint}{\constraint{Disjoint}\xspace}
\newcommand{\common}{\constraint{Common}\xspace}
\newcommand{\softalldiff}{\constraint{SoftAllDifferent}\xspace}
\newcommand{\notallequal}{\constraint{Not-All-Equal}\xspace}
\newcommand{\softallequal}{\constraint{Soft-All-Equal}}
\newcommand{\gsc}{\constraint{GSC}}
\newcommand{\sequence}{\constraint{Sequence}\xspace}
\newcommand{\precedence}{\constraint{Precedence}\xspace}
\newcommand{\GCC}{\constraint{GCC}\xspace}
\newcommand{\roots}{\constraint{Roots}\xspace}
\newcommand{\range}{\constraint{Range}\xspace}

\newcommand{\tighter}{\mbox{$\preceq$}}
\newcommand{\stighter}{\mbox{$\prec$}}
\newcommand{\incomparable}{\mbox{$\bowtie$}}
\newcommand{\equivalent}{\mbox{$\equiv$}}

\newcommand{\todo}[1]{{\tt (... #1 ...)}}
\newcommand{\myOmit}[1]{}

\title{Is Computational Complexity a Barrier to Manipulation?}

\author{Toby Walsh}
\institute{NICTA and University of NSW,
Sydney, Australia, email: 
toby.walsh@nicta.com.au}

\pdfinfo{
 /Author (Toby Walsh)
  /Title (Is Computational Complexity a Barrier to Manipulation?)
  /Keywords (social choice, manipulation, computational complexity,
  stable marriage)
  /DOCINFO pdfmark}

%\date{1st April 2009}

\maketitle

\begin{abstract}
When agents are acting together,
they may need a simple mechanism
to decide on joint actions. 
One possibility is to have
the agents express their preferences
in the form of a ballot and
use a voting rule to decide
the winning action(s). 
Unfortunately, agents may try to manipulate
such an election by mis-reporting their preferences. 
Fortunately, it has been shown that it is NP-hard
to compute how to manipulate a number of 
different voting rules. However, NP-hardness only
bounds the worst-case complexity. Recent theoretical
results suggest that manipulation may often
be easy in practice. To address this issue, I suggest
studying empirically if computational complexity is in
practice a barrier to manipulation. The basic tool used in my
investigations is the identification of computational
``phase transitions''. Such an approach has been
fruitful in identifying hard instances of
propositional satisfiability and other NP-hard problems. 
I show that phase transition behaviour gives
insight into the hardness of manipulating 
voting rules, increasing concern that
computational complexity is indeed any sort of barrier. 
Finally, I look at the problem of computing
manipulation of other, related problems like
stable marriage and tournament problems. 
\end{abstract}

\section{Introduction}

The Gibbard Satterthwaite theorem proves that, under some
simple assumptions, a voting rule can always be manipulated. 
%Such manipulation of a voting
%rule is generally considered undesirable. 
In an influential paper \cite{bartholditoveytrick},
Bartholdi, Tovey and Trick 
proposed an appealing escape:
perhaps it is computationally so difficult to find a successful
manipulation that agents have little
option but to report their true preferences? 
To illustrate this idea, 
they demonstrated that the second order Copeland rule
is NP-hard to manipulate. 
Shortly after, Bartholdi and Orlin proved
that the more well known Single Transferable Voting
(STV) rule is NP-hard to manipulate \cite{stvhard}.
Many other voting rules have subsequently
been proven to be NP-hard
to manipulate \cite{csljacm07}.
There is, however, increasing concern that worst-case results
like these do not reflect the difficulty of manipulation
in practice. Indeed, several theoretical results 
suggest that manipulation may often be 
easy (e.g. \cite{xcec08b}). 

\section{Empirical analysis}

In addition to
attacking this question theoretically, I have argued in 
a recent series of papers that
we may benefit from studying it empirically \cite{wijcai09,wecai10}. 
There are several reasons why empirical
analysis is useful. 
First, theoretical analysis is often
restricted to particular distributions like uniform votes. 
Manipulation may be very different in practice due to 
correlations in the preferences of the agents. For instance, 
if all preferences are single-peaked then there
voting rules where it is in the best interests of all agents to state
their true preferences. Second, theoretical 
analysis is often asymptotic so does not
reveal the size of hidden constants. The size of such
constants may be important to 
the actual computational cost of computing
a manipulation. 
In addition, elections are typically bounded in
size. Is asymptotic behaviour
relevant to the size of elections met in practice? 
An empirical study may quickly suggest if the result
extends to more candidates. 
Finally, empirical studies can suggest theorems
to prove. For instance, our experiments
suggest a simple formula
for the probability that a coalition
is able to elect a desired candidate. 
It would be interesting to derive 
this exactly. 

\section{Voting rules}

My empirical studies have focused on two voting rules: single transferable voting (STV)
and veto voting. STV is representative of voting rules
that are NP-hard to manipulate without weights on votes.
Indeed, as I argue shortly, it is one of the few such rules.
Veto voting is, on the other hand, a simple representative
of rules where manipulation is
NP-hard when votes are weighted or (equivalently) 
we have uncertainty about how agents have
voted. The two voting rules therefore cover the
two different cases where computational complexity
has been proposed as a barrier to manipulation. 

STV proceeds in a number of rounds.
%STV is then equivalent to the instant-runoff voting rule.
%STV can, however, be used in elections with multiple
%winners. 
Each agent totally ranks the candidates on a ballot. 
Until one candidate has a majority of
first place votes, we eliminate the candidate
with the least number of first place votes
Ballots placing the eliminated candidate in first 
place are then re-assigned to the second place candidate.
STV is used in a wide variety of elections
including for the
Australian House of Representatives, 
the Academy awards,
and many organizations including the 
American Political Science Association,
and the International Olympic Committee.
STV has played a central role in the
study of the computational complexity of manipulation.
Bartholdi and Orlin argued that:
\begin{quote}
{\em ``STV is apparently
unique among voting schemes in actual use today 
in that it is computationally
resistant to manipulation.'' }
(page 341 of \cite{stvhard}).
\end{quote}
%Whilst there exist other voting rules which are NP-hard to manipulate,
%computational complexity is either restricted to 
%less well known rules like second order
%Copeland or to more well known voting rules but with
%the restriction that there are large
%weights on the votes. STV is the only commonly used voting rule
%that is NP-hard to manipulate without weights
%\cite{stvhard}. 

By comparison, the veto rule is a much simpler scoring rule
in which each agent gets to cast a veto against one candidate. 
The candidate with the fewest vetoes wins. 
There are several
reasons why 
the veto rule is interesting to study. The veto rule is very
simple to reason about. This can be contrasted
with other voting rules like STV. Part of the 
complexity of manipulating the STV rule
appears to come from reasoning about 
what happens between the different rounds. 
The veto rule, on the other hand, has 
a single round. 
The veto rule is also on the borderline of tractability
since constructive manipulation (that is, ensuring 
a particular candidate wins) of the veto rule by a coalition of
weighted agents is NP-hard but destructive manipulation (that is,
ensuring a particular candidate does not win) is polynomial \cite{csljacm07}.

\section{Voting distributions}

Empirical analysis requires collections of votes
on which to compute manipulations. 
My analysis starts with one of the simplest possible
scenarios: elections in which each vote is equally
likely. We have one agent trying to manipulate an
election of $m$ candidates in which $n$ other
agents vote. Votes are drawn uniformly
at random from all $m!$ possible votes. 
This is the Impartial Culture (IC) model. 
In many real life situations, however, votes are 
correlated with each other. I therefore also 
considered single-peaked preferences, single-troughed
preferences, and votes
drawn from the Polya Eggenberger urn model \cite{polya-urn}.
In an urn model, 
we have an urn containing all possible votes. 
We draw votes out of the urn at random, and
put them back into the urn with $a$ 
additional votes of the same type (where
$a$ is a parameter).
%As $a$ increases, there is 
%increasing correlation between the votes. 
This generalizes both the Impartial Culture
model ($a=0$) and the Impartial Anonymous Culture ($a=1$) model. 
Real world elections may differ
from these ensembles. 
I therefore also sampled some real voting
records \cite{isai95,ghpwaaai99}. Finally, one 
agent on their own is often unable to manipulate
the result. I therefore also
considered coalitions of agents who are trying
to manipulate elections. 

\section{Results}

My experiments suggest
different behaviour occurs in the problem of 
computing manipulations
of voting rules
than in other NP-hard problems like
propositional satisfiability and graph colouring 
\cite{cheeseman-hard,waaai98}. 
For instance, we often did not see a rapid transition
that sharpens around a fixed point
as in satisfiability \cite{mitchell-hard-easy}. 
Many transitions 
appear smooth and do not sharpen 
towards a step function as problem size increases. 
Such smooth phase transitions have been
previously seen in polynomial problems 
\cite{waaai2002}.
In addition, hard instances often did not
occur around some critical parameter.
Figures 1 to 3 reproduce some typical
graphs from \cite{wecai10}.

\begin{figure}[hptb]
\vspace{-2.5in}
\begin{center}
\includegraphics[scale=0.4]{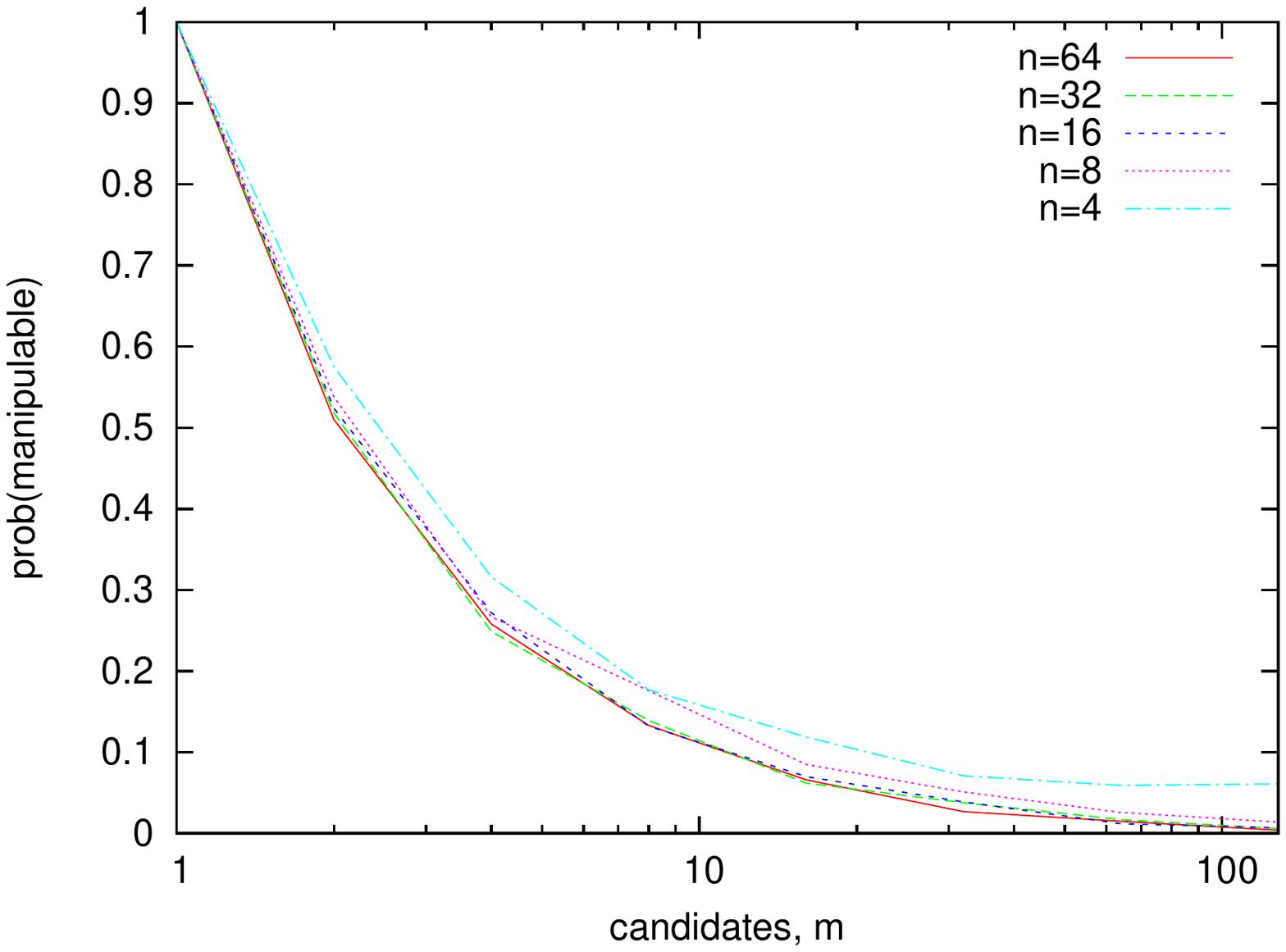}
\end{center}
\vspace{-0.5in}
\caption{Manipulability of 
correlated votes. %}
%{\textnormal 
The number of agents $n$
is fixed and we vary the number of
candidates $m$. 
The y-axis measures the
probability that the manipulator can make a random
candidate win.
}

\label{fig-urn-prob-varm}
\end{figure}
\begin{figure}[hptb]
\vspace{-2.5in}
\begin{center}
\includegraphics[scale=0.4]{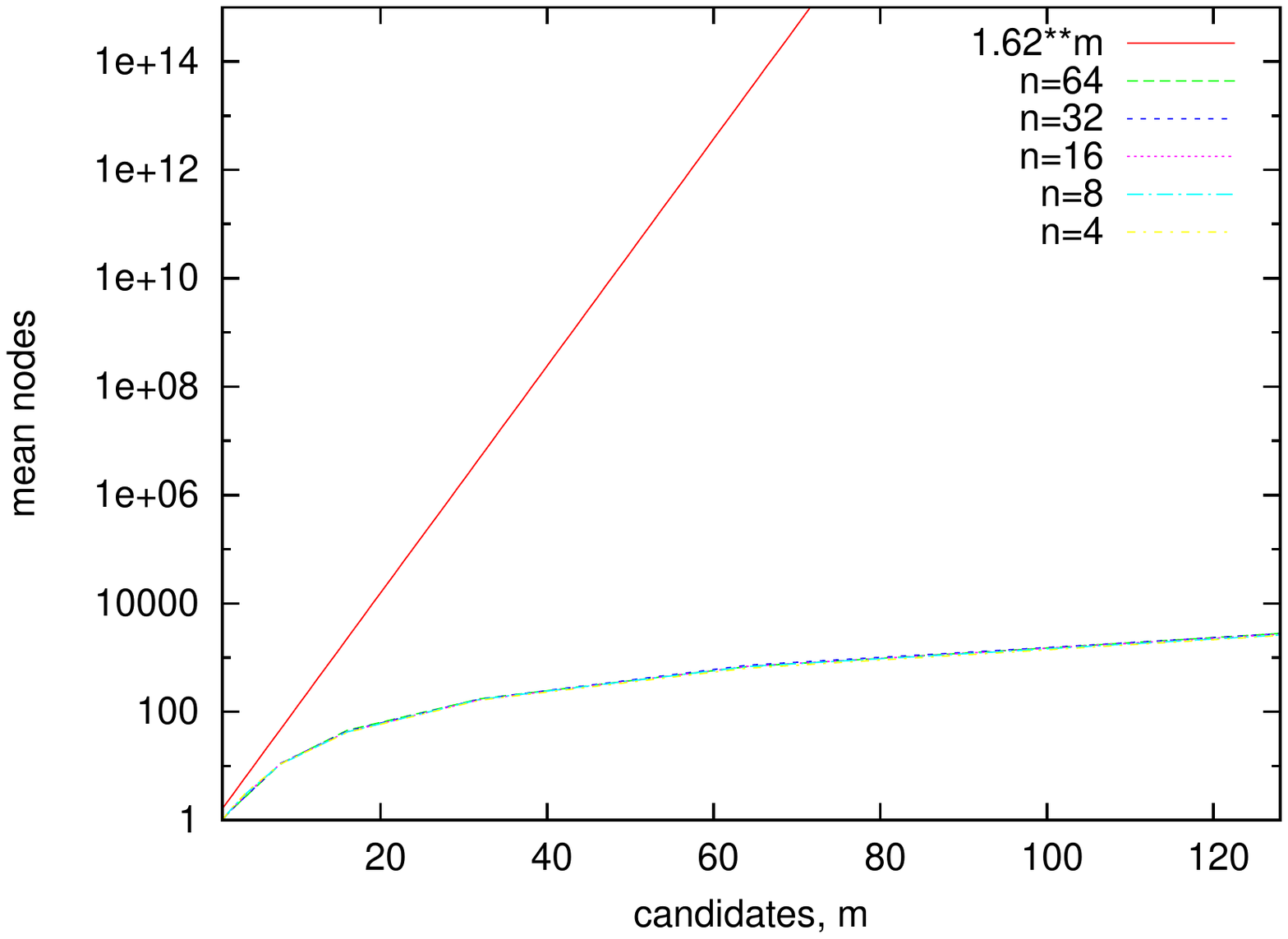}
\end{center}
\vspace{-0.5in}
\caption{Search to compute if an agent can manipulate an election
with correlated votes. %}
%{
The number of agents $n$
is fixed and we vary the number of
candidates $m$. 
%The $n$ fixed votes are drawn using the Polya Eggenberger urn  model
%with $b=1$. 
The y-axis measures the mean 
number of search nodes explored to compute
a manipulation or prove that none exists. 
Median
and other percentiles are similar.
$1.62^{m}$ is the published worst-case
bound for the recursive algorithm
used to compute a manipulation \cite{csljacm07}.
}
\label{fig-urn-nodes-varm}
\end{figure}

\begin{figure}[hptb]
\vspace{-2.5in}
\begin{center}
\includegraphics[scale=0.4]{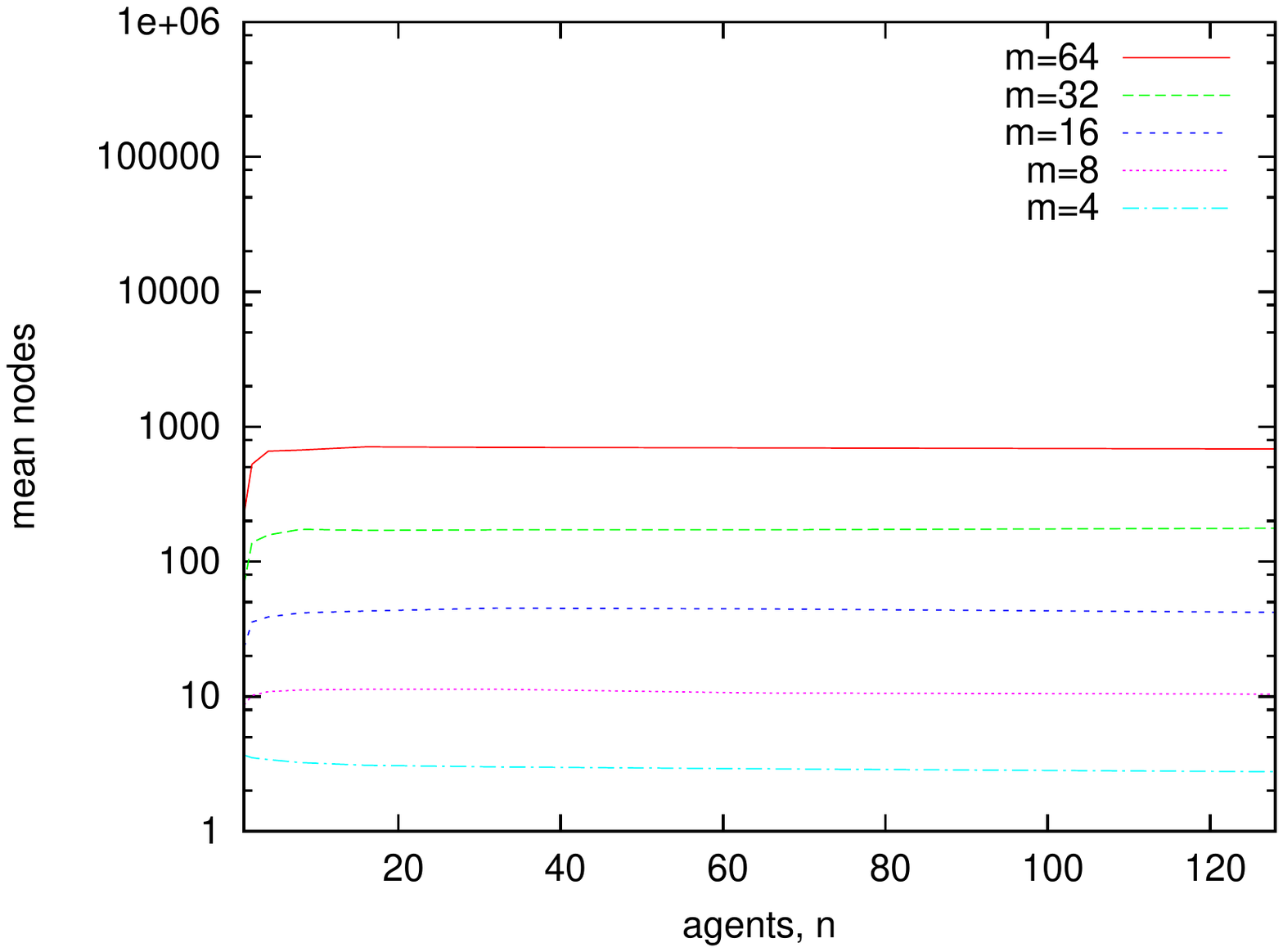}
\end{center}
\vspace{-0.5in}
\caption{Search to compute if an agent can manipulate an election
with correlated votes. %}
%{
The number of candidates $m$ is fixed and we vary the number of
agents $n$. % voting. 
%The y-axis measures the mean 
%number of search nodes explored to compute 
%a manipulation or prove that none exists. Median
%and other percentiles are similar.
}
\label{fig-urn-nodes-varn}
\end{figure}

Similar phase transition studies have been used to
identify hard instances of NP-hard problems
like propositional
satisfiability  \cite{mitchell-hard-easy,SAT-phase,pub702},
constraint satisfaction \cite{Prosser94,secai94,gmpwcp95,gmpwaaai96,random},
number partitioning \cite{mertensnp,rnp,gw-ci98}, Hamiltonian
circuit \cite{fgwipl,vandegriend-culberson}, 
and the traveling salesperson problem \cite{GentIP:tsppt,zhangjair2004}.
Phase transition studies have also been
used to study polynomial problems
\cite{gsecai96,gmpswcp97}
as well as higher complexity
classes \cite{gwaaai99,bailey1}
and optimization problems
\cite{swsat2002,larrosa2}.
Finally, phase transition studies have
been used to study problem structure
like small worldiness \cite{wijcai99}
and high degree nodes \cite{wijcai2001}.

\section{Other manipulation problems}

Another multi-agent problem in which manipulation
may be an issue is the stable marriage problem.
This is the well-known problem
of matching men to women so that no man and woman who
are not married to each other both prefer each other. 
It has a wide variety of
practical applications such as a matching 
doctors to hospitals. 
%As a practical
%example, the US Navy has a web-based multi-agent system
%for assigning sailors to ships \cite{liebowitz}.
As with voting, an important issue is whether agents 
can manipulate the result by
mis-reporting their preferences. 
Unfortunately, Roth \cite{roth-manip} proved 
that {\em all} stable marriage procedures
can be manipulated. 
We might hope that 
computational complexity might also be a barrier
to manipulate stable marriage procedures. 
In joint work with
Pini, Rossi and Venable, I have proposed a new stable marriage procedures that
is NP-hard to manipulate \cite{prvwaamas09}. 
Another advantage of this new procedure is that, unlike
the Gale-Shapley algorithm, it does not favour one
sex over the other. 
Our procedure picks the stable matching that is most preferred
by the most popular men and women. 
The most preferred men and women are chosen
using a voting rule. We prove that, if the voting rule used is STV
then the resulting stable matching procedure is also 
NP-hard to manipulate.
We conjecture that other voting rules which are NP-hard
to manipulate will give rise to stable matching procedures
which are also NP-hard to manipulate. 

%\section{Tournament manipulation}

The final domain in which I have studied 
computational issues surrounding manipulation 
is that of (sporting) tournaments (joint work
with Russell) \cite{rwadt09}. 
Manipulating a tournament is slightly
different to manipulating an election. In a sporting
tournament, the voters are also the candidates. 
Since it is hard (without bribery or similar mechanisms)
for a team to play better than it can, we consider
just manipulations where the manipulators can
throw games. %By comparison, in an election, voters in
%the manipulating coalition can mis-report their
%preferences in any way they choose. 
We show that we can
decide how to manipulate round robin and cup competitions, two 
of the most popular sporting competitions 
in polynomial time. In addition, we show that finding the minimal 
number of games that need to be thrown to manipulate the
result can also be determined in polynomial time.  
Finally, we give a polynomial time
proceure to calculate the probability that a team wins a cup competition 
under manipulation. 
%These results might be of interest to people
%who follow sporting tournaments. 

%\section{Related work}

\section{Conclusions}

I have argued that empirical studies can
provide insight into 
whether computational
complexity is a barrier to the manipulation.
Somewhat surprisingly, almost every one of the 
many millions of elections in the
experiments in \cite{wijcai09,wecai10}
was easy to manipulate or to prove
could not be manipulated. 
Such experimental results increase the concerns that
computational complexity is indeed a barrier
to manipulation in practice. 
Many other voting rules have been
proposed which could be studied in the future. Two interesting
rules are maximin and ranked pairs. These
two rules have only recently been shown to be NP-hard
to manipulate, and
are members of the small set of voting rules which 
are NP-hard to manipulate without weights
or uncertainty 
\cite{xzpcrijcai09}. 
These results demonstrate that
empirical studies can provide insight into the computational
complexity of computing manipulations. 
It would be interesting to consider
similar phase transition studies for related problems like
preference elicitation \cite{waaai2007,waamas08}.

\subsubsection*{Acknowledgements:}

NICTA is funded by 
the Department of Broadband, 
Communications and the Digital Economy, and the 
Australian Research Council. 

\bibliographystyle{splncs}

%\bibliographystyle{aaai}
%%\bibliographystyle{alpha}
%%\bibliography{/home/s5/tw/biblio/a-z,/home/s5/tw/biblio/pub}
%\bibliography{/home/tw/biblio/a-z,/home/tw/biblio/pub,/home/tw/biblio/a-z2,/home/tw/biblio/pub2}
%%\bibliography{/n/endjinn/u6/tw/biblio/a-z,/n/endjinn/u6/tw/biblio/pub}
%%\bibliography{/usr/tw/biblio/a-z,/usr/tw/biblio/pub}
%\bibliography{/Users/twalsh/Documents/biblio/a-z,/Users/twalsh/Documents/biblio/pub,/Users/twalsh/Documents/biblio/a-z2,/Users/twalsh/Documents/biblio/pub2}
%%\bibliography{/home/arp/disk1/tw/biblio/a-z,/home/arp/disk1/tw/biblio/pub}
%%\bibliography{/u6/tw/biblio/a-z,/u6/tw/biblio/pub}
%\bibliography{/usr/local/users/tw/biblio/a-z,/usr/local/users/tw/biblio/pub}
%\bibliography{biblio}

%\end{thebibliography}

\end{document}